# A Multi-Criteria Evaluation of Color Normalization Methods in Histopathology: Nuclei Preservation, Artifact Sensitivity, Background Fidelity, and Downstream Performance

Falah Jabar[1,2]*, Aria Torkpour[2], Pasquale Lombardi[2], Masoud Tafavvoghi[1], Per Niklas Benzler Waaler[3], Sigve Andersen[4,5], Ilona Urbarova[4,6], David J. Pinato[2,7], Mehrdad Rakaee[8], Tom Donnem[4,9], Lill-Tove Rasmussen Busund[1,3], Elin Richardsen[1]

[1] Dep. of Clinical Pathology, University Hospital of North Norway, Tromsø, Norway.
[2] Dep. of Surgery and Cancer, Imperial College London, London, United Kingdom.
[3] Dep. of Medical Biology, UiT The Arctic University of Norway, Tromsø, Norway.
[4] Dep. of Clinical Medicine, UiT The Arctic University of Norway, Tromsø, Norway.
[5] Dep. of Cancer, University Hospital of North Norway, Tromsø, Norway.
[6] Dep. of Community Medicine, UiT The Arctic University of Norway, Tromsø, Norway.
[7] Dep. of Translational Medicine, University of Piemonte Orientale, Novara, Italy.
[8] Dep. of Cancer Genetics, Oslo University Hospital, Oslo, Norway.
[9] Dep. of Oncology, University Hospital of North Norway, Tromsø, Norway.

*Corresponding author:
Falah Jabar, PhD
Hansine Hansens veg 67, 9019 Tromsø, Norway
E-mail: f.rahim@imperial.ac.uk

**Abstract**—Deep learning models for nuclei detection and classification may generalize poorly across medical centers because of variations in staining and scanning. Although color normalization aims to reduce these variations, its effects on nuclear structure, background preservation, artifact sensitivity, and nuclei detection and classification performance have not been systematically evaluated. We present a procedure to benchmark 11 classical and deep learning normalization methods using 7 public and private nuclei datasets. Additionally, we investigate the effect of 7 input color spaces. The normalization methods differed in their effects on nuclear structure, background, and sensitivity to artifacts. However, detection and classification performance varied more across datasets than across normalization methods, and the input color space had limited influence. No single method performed best on all criteria, and image-level metrics did not reliably predict downstream performance, highlighting the need for nuclei- and structure-aware normalization methods. To support this work, we introduce WSI Quality Control and Normalization (WSI-QCN), a publicly available pipeline for robust whole-slide image quality assessment and color normalization; the source code is available on GitHub.

## 1. Introduction

Whole-slide images (WSIs) are gigapixel digital representations of thin tissue sections obtained using high-resolution slide scanners. However, WSIs often show considerable inter-laboratory color variation due to differences in staining protocols and scanner characteristics (Vahadane *et al.*, 2016; Khan *et al.*, 2026). This variability poses a significant challenge to the application of deep learning (DL) in computational pathology, as models trained on WSIs from one laboratory may fail when applied to WSIs from other laboratories, limiting their generalizability (Tellez *et al.*, 2019; Lin *et al.*, 2025; Khan *et al.*, 2026).

Color normalization is a widely used approach to reduce the effects of color variation. A range of normalization techniques has been proposed, spanning traditional methods that rely on statistical analysis and stain separation to more recent approaches that leverage DL models (see Table S1 in the supplementary materials). Fig. 1 illustrates four representative hematoxylin and eosin (H&E)-stained image tiles normalized using three popular methods: Macenko (Macenko *et al.*, 2009), CycleGan (Zhu *et al.*, 2017), and StainNet (Kang *et al.*, 2021). Although these methods can reduce color variation, they may introduce: (1) visual artifacts and changes in staining intensity, (2) alterations in tissue and nuclei morphology, (3) texture or color to background, and (4) inconsistent normalization in the presence of tissue artifacts. These changes can affect nuclei detection and classification by varying nuclei features. Notably, some normalization methods rely on H&E stain separation, distinguishing H, which primarily stains cell nuclei, from E, which primarily stains the cytoplasm and other tissue structures. Any changes in H&E and reconstruction may therefore change nuclei appearance.

A few studies have benchmarked color normalization methods (Tellez *et al.*, 2019; Xu *et al.*, 2025; Hoque *et al.*, 2024; Khan *et al.*, 2026; Lin *et al.*, 2025), primarily evaluating their performance in tile-level classification and slide-level prognosis. However, the effects of color normalization on nuclei morphology and appearance, and its impact on nuclei detection and classification performance, remain unexplored. Furthermore, the behavior of normalization methods on background regions, as well as their sensitivity to artifacts such as tissue folds and blurring, has not been investigated. To address these gaps, this study presents a nuclei-aware evaluation of color normalization methods, examining their effects on nuclei detection and classification, nuclei appearance, background preservation, and artifact sensitivity.

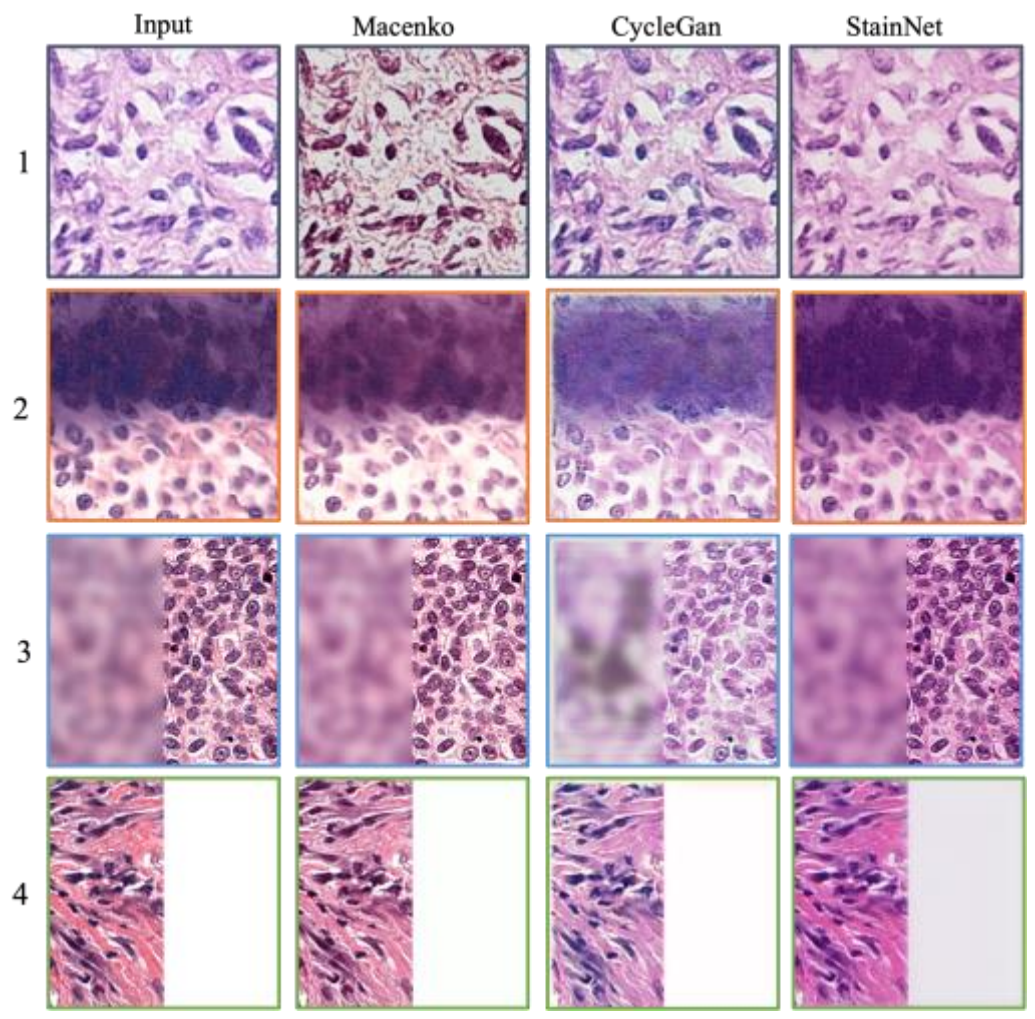

**Fig. 1**. Visual comparison of four H&E-stained image tiles before and after normalization using Macenko, CycleGAN, and StainNet, highlighting differences in color, tissue and nuclear structure, and background preservation.

The main contributions of this study are summarized as follows:

- Impact of color normalization on nuclei detection and classification: A comparative evaluation of how different normalization methods affect nuclei detection and classification performance across 6 multicenter datasets.
- Nuclei-aware evaluation of color normalization: A systematic evaluation of color normalization methods across 7 multicenter nuclei datasets, focusing on changes in nuclei appearance, morphology, and structural characteristics.
- Background preservation analysis: An evaluation of how effectively normalization methods preserve clean, uniform backgrounds without introducing color variations or texture.
- Artifact sensitivity analysis: An investigation of color normalization robustness to image artifacts (e.g., folds and blurring) and how these artifacts affect normalization results.
- Impact of input color domain on nuclei detection and classification: An investigation of how different color representations influence nuclei detection and classification performance across the same 6 multicenter datasets.
- Development of WSI Quality Control and Normalization (WSI-QCN), a publicly available pipeline for robust whole-slide image quality assessment and color normalization.

The rest of this paper is organized as follows: Section 2 reviews related work; Section 3 describes the datasets used; Section 4 describes the methodology with technical details; Section 5 presents the experimental results; Section 6 describes the WSI quality control and color normalization pipeline; Section 7 concludes the paper with future work directions.

## 2. Related Work

Color normalization methods can be divided into traditional and deep learning-based approaches (Table S1 in the supplementary materials):

- **Traditional approaches**–Ruifrok *et al*., 2001; Macenko *et al*., 2009; Ivanov *et al*., 2025; Vahadane *et al*., 2016; Khan *et al*., 2014 are stain separation-based methods. These methods differ in how the stain matrix is obtained and in the number of reference images used. Each image is converted to optical density (OD) and decomposed with a stain matrix into hematoxylin and eosin concentrations. These concentrations are rescaled to match the reference image, and the image is reconstructed with the reference stain matrix. Reinhard *et al.*, 2001 and histogram matching (Gonzalez *et al.*, 2018) are global color-statistics methods that match the source image's overall color distribution to the reference. Traditional methods require no training, but typically rely on one or a few reference images, making normalization sensitive to reference selection. They may fail under stain variation, overlap, saturation, or dominant stains, and generally ignore spatial structure.

- **Deep learning approaches**–CycleGan (Zhu *et al.*, 2017), StainGan (Shaban *et al.*, 2019), StainNet (Kang *et al.*, 2021), pix2pix (Isola *et al.*, 2017), SastainDiff (Yang *et al.*, 2025) perform normalization as image-to-image translation from data rather than from a single reference image. Unlike traditional methods that rely on explicit stain decomposition or reference-image statistics, these approaches can learn complex, nonlinear relationships in stain appearance and accommodate variations in staining patterns. However, they require representative training data and computational resources, and their performance depends on the training data, source and target domains, and model architecture. Furthermore, image translation may introduce artifacts, vary nuclei morphology, or generate misleading structures, potentially affecting downstream nuclei detection and classification performance (Xu *et al.*, 2025; Khan *et al.*, 2026).

## 3. Data Material

Seven public and private multicenter H&E-stained nuclei datasets were used: CoNSeP (Graham *et al*., 2019), Lizard (Graham *et al*., 2021), NuCLS (Amgad *et al*., 2022), PanopTILs (Liu *et al*., 2024), SegPath (Komura *et al*., 2023), PanNuke (Jevgenij *et al*., 2019), and ILCD (Falah *et al*., 2026). These datasets differ in tissue type, scanner, magnification, annotation procedure, cell categories, and tile size. As described in Jabar *et al.*, 2026, all datasets were preprocessed to standardize image tiles to 256 × 256 pixels, keeping the original magnification. Furthermore, the original cell labels were merged into three common categories: immune cells, tumor cells, and other cells. This standardization enabled consistent comparisons across datasets. To investigate the effects of artifacts and background on color normalization methods, an additional multi-institutional artifact dataset introduced in (Jabar *et al.*, 2025) was used. This dataset includes artifact-free tiles and tiles containing folding and blurring artifacts. Detailed descriptions of all datasets are provided in the supplementary material.

## 4. Method

This section describes the experimental design and evaluation procedure. First, we assessed how color normalization affects nuclei appearance, background preservation, and sensitivity to artifacts. Then, we evaluated the impact of color normalization and input color space on nuclei detection and classification (**Fig. 2**).

### *4.1. Evaluation of Nuclei Structure Preservation–Fig. 2a*

Color normalization methods may change nuclei appearance but should preserve nuclei structure. To assess this, each normalized tile was compared with its original counterpart within the ground-truth nuclei regions, across all 7 datasets: PanNuke, CoNSeP, ILCD, Lizard, NuCLS, PanopTILs and SegPath, and all 11 normalization methods. For each tile, nuclei pixels were extracted from the nuclei mask (mask > 0), and the same pixels were used in both images, restricting the comparison to nuclei regions. The primary metric was the structural similarity index (SSIM), computed between the original and normalized images and averaged over nuclei pixels. In addition, the CIEDE2000 color difference (ΔE00), Jensen–Shannon divergence (JSD), histogram intersection (HI), and Pearson correlation (PC) were also computed (Xu *et al*., 2025; Khan *et al*., 2026; *Xu et al*., 2025; Hoque *et al*., 2024). These color metrics describe how strongly each method modifies nuclei color. All metrics were computed per tile and averaged for each dataset.

### *4.2. Background Preservation–Fig. 2b*

Color normalization methods should normalize tissue regions without painting non-tissue regions (background). To test this, the artifact dataset (Section 3), which contains two common artifacts, tissue folds and blurring, each with a pixel-level annotation mask (0 = background, 1 = valid tissue, 2 and 3 = artifact), was used. Pixels labeled as artifacts were replaced with near-white (RGB 250, 250, 250), creating a known background region whose correct appearance after normalization is near-white. Each modified tile was then normalized with each of the 11 methods. Evaluation was restricted to the whitened region, which was eroded by 2 pixels to exclude boundary pixels affected by edge blending. Three metrics were computed within the whitened region, comparing each normalized image with its whitened input. Background ΔE00, quantifying the overall color shift. Painted background percentage (PBP), which is the proportion of pixels with ΔE00 > 5, quantifying how much of the background was visibly painted. Background gradient (BG), which is the mean gradient magnitude in the normalized image, quantifying texture introduced into a region that should be uniform. Metrics were averaged over all tiles.

### *4.3. Artifact Sensitivity–Fig. 2c*

Traditional color normalization methods estimate their parameters from the entire tile, so artifacts present in the tile can bias the parameter values. To quantify this effect, each tile in the artifact dataset (Section 3) was processed in two versions. In the first, the original tile, including the artifact, was normalized directly. In the second, artifact pixels were replaced with near-white values before normalization. Both versions were normalized with the same method and settings, and this procedure was repeated for each of the 11 methods. Assuming the tissue outside the artifact is identical in both inputs, any difference between the two normalized outputs in that region can be attributed to the artifact's influence on the normalization. Comparisons were restricted to nuclei pixels within valid tissue, defined as pixels that belong to a nucleus in the nuclei mask and are labeled as valid tissue in the artifact mask. Artifact regions and background were therefore excluded. Five metrics were computed between the two normalized outputs: SSIM, ΔE00, JSD, HI, and PC, averaged over all tiles.

### *4.4. Impact of Color Normalization on Nuclei Detection and Classification–Fig. 2d*

To evaluate the effects of color normalization on nuclei detection and classification, CP-Net (Jabar *et al*., 2026)–a nuclei detection and classification pipeline–was trained and evaluated using each of the 11 normalization methods. In a benchmark against ten state-of-the-art pipelines across eight datasets, CP-Net achieved the most favorable trade-off between accuracy and inference time (Jabar *et al*., 2026). Its relatively lightweight architecture also makes it practical to train separate models for each normalization method. For each method, the PanNuke training set was normalized, and a separate CP-Net model was trained, resulting in 11 models. All model hyperparameters, GPU hardware, and training configurations followed the original CP-Net study (Jabar *et al*., 2026) to ensure consistency across experiments. Each model was then evaluated on six external test sets: CoNSeP, ILCD, Lizard, NuCLS, PanopTILs, and SegPath, with each test set normalized using the same method as the corresponding training set. As none of the test sets were used during training, this design enabled an assessment of how each normalization method affected generalization to unseen datasets. Predictions were compared with ground-truth annotations using segmentation quality (SQ), detection quality (DQ), and panoptic quality (PQ), yielding 66 model–dataset evaluation combinations.

### *4.5. Impact of Input Color Space–Fig. 2e*

The impact of input color space on deep learning performance remains understudied in computational pathology, where RGB is used as the input color space. The choice of input color space has been shown to influence deep learning performance in histological colon tissue classification (Velastegui *et al*., 2021). Whether similar effects hold for nuclei detection and classification remains unknown. To investigate this, CP-Net was trained and evaluated using seven input color spaces: RGB as the reference, HSV, HLS, YCbCr, XYZ, LAB, and LUV. A separate CP-Net (Jabar *et al.*, 2026) model was trained for each input color space using the PanNuke training set, resulting in 7 models. Each model was evaluated on six external test datasets: CoNSeP, ILCD, Lizard, NuCLS, PanopTILs, and SegPath. Before inference, each test tile was converted to the input color space used to train the corresponding model. Performance was assessed using DQ, SQ and PQ, expressed relative to the RGB model on the same test dataset: $\Delta M(c, d) = M(c, d) - M(RGB, d)$, where $M \in \{DQ, SQ, PQ\}$, c denotes the input color space, and d the test dataset. Positive values indicate a gain over RGB, and negative values indicate that RGB performed better.

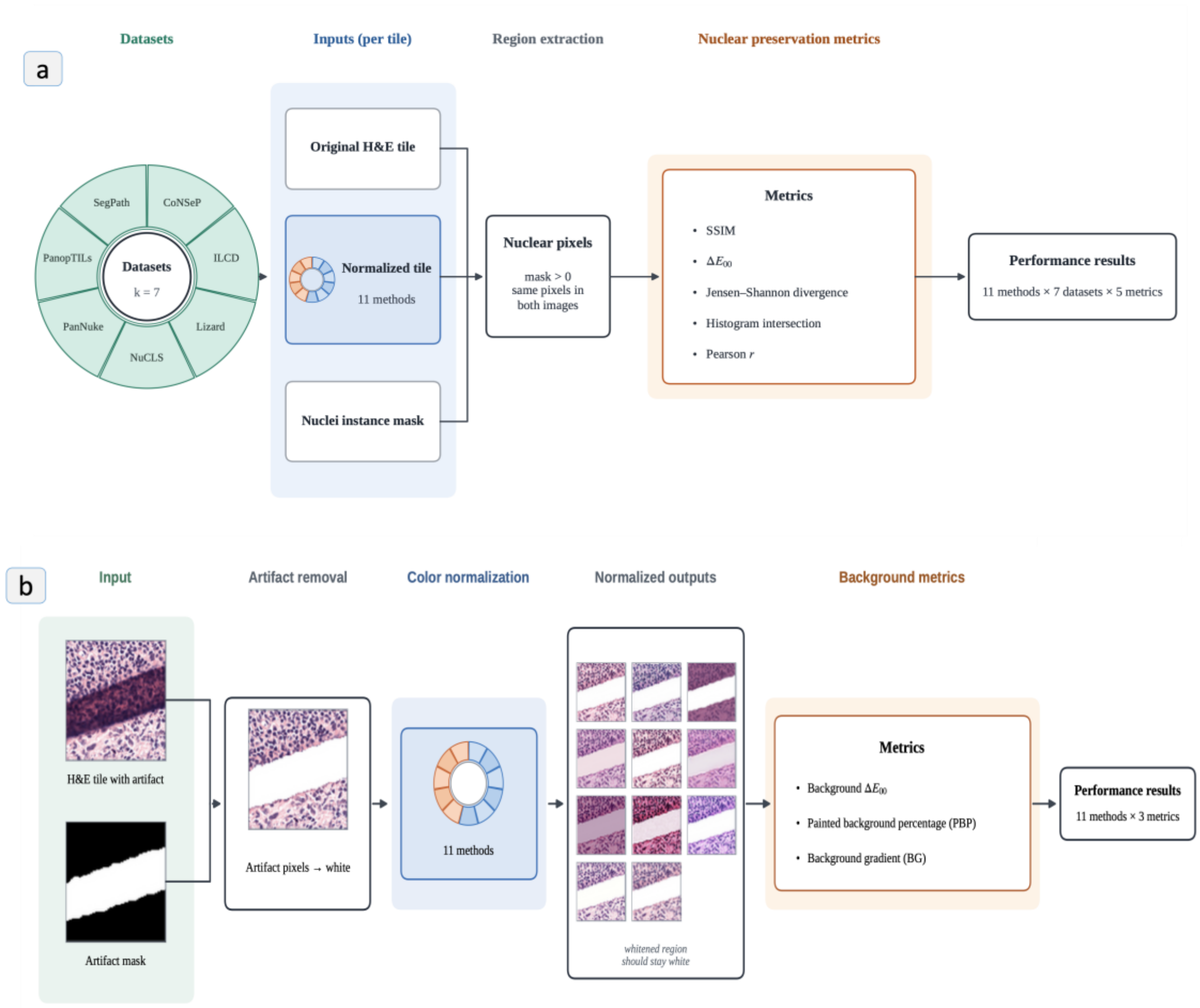

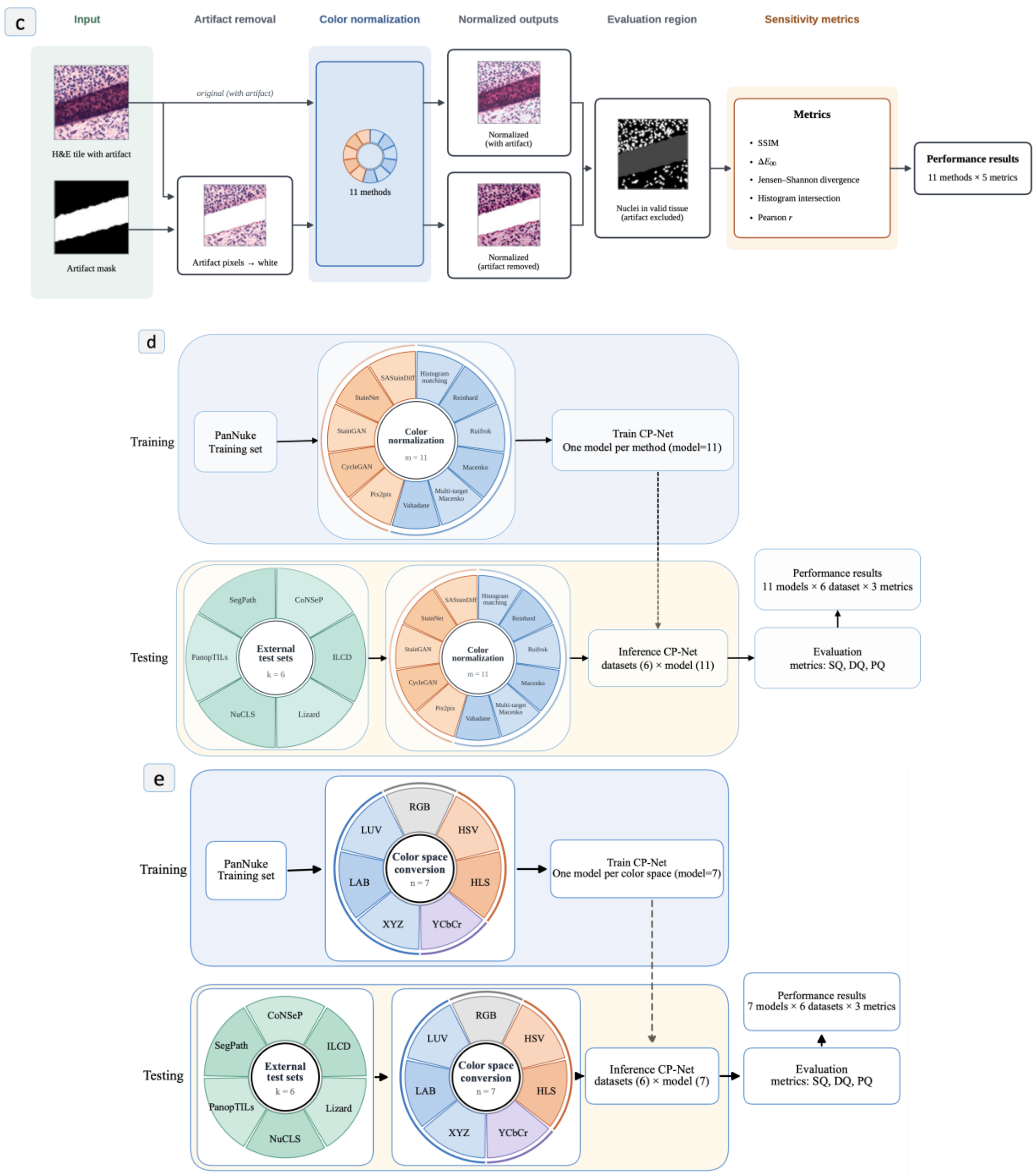

**Fig. 2. Experimental design overview**: (a) Evaluation of nuclei structure–original and normalized tiles are compared within nuclei pixels; (b) Background preservation–artifact pixels are set to near-white before normalization, and the whitened region is assessed; (c) Artifact sensitivity–normalized outputs with and without the artifact are compared over nuclei pixels in tissue regions; (d) Impact of color normalization–one CP-Net model is trained per method on PanNuke and evaluated on six external test sets; (e) Impact of input color space–one CP-Net model is trained per input color space on PanNuke and evaluated on six external test sets.

## 5. Performance Evaluation

This section presents quantitative and qualitative evaluations of the 11 color normalization methods in terms of nuclear structure preservation, background preservation, artifact sensitivity, nuclei detection and classification performance, input color space, and computational cost.

### *5.1. Evaluation of Nuclei Structure Preservation*

Fig. 3a shows the metric values for each color normalization method, averaged across the 7 nuclei datasets. Nuclei structural preservation varied considerably across methods, with SSIM values ranging from 0.75 to 0.99. StainNet (0.99), Pix2Pix (0.98), and StainGAN (0.94) achieved the highest SSIM values. Multi-Macenko, CycleGAN, Vahadane, Ruifrok, Macenko showed intermediate performance (0.88–0.91), while Histogram Matching (0.85), SAStainDiff (0.81), and Reinhard (0.75) obtained the lowest SSIM values. The color metrics Pix2Pix, CycleGAN, and StainNet produced the smallest change in nuclei color (ΔE00 0.59–0.64, JSD 0.30–0.32, HI 0.62–0.66, PC 0.69–0.72), whereas Ruifrok produced the largest (ΔE00 0.99, JSD 0.51, HI 0.39, PC 0.33), followed by SAStainDiff and Reinhard. Notably, the magnitude of color change did not determine nuclei structural preservation. Ruifrok produced the largest color change while maintaining structural similarity comparable to other methods (SSIM 0.91). In contrast, CycleGAN introduced only modest color changes but ranked fifth in structural similarity, also with an SSIM of 0.91. Thus, color modification and loss of nuclei structure can be independent properties of normalization methods, and structural similarity must be assessed separately from color change. A qualitative comparison of the methods is provided in supplementary materials Fig. S1. Metric values for each dataset are provided in supplementary materials Fig. S4.

### *5.2. Background Preservation*

Methods differed substantially in their preservation of white background regions (Fig. 3b). Histogram Matching, Vahadane, Ruifrok, and Macenko strongly preserved the white background ($\Delta E_{00} \leq 0.74$, PBP ≤ 0.16%, BG ≈ 0). Multi-Macenko and CycleGAN introduced minor colour shifts ($\Delta E_{00} = 2.49$ and 0.99, respectively). The remaining methods change the background more strongly. Reinhard and StainNet introduced uniform colour change across the entire background (PBP = 100%; $\Delta E_{00}$ = 27.68 and 11.39, respectively) without introducing additional texture (BG ≈ 0), consistent with pixel-wise colour transformations. In contrast, Pix2Pix, StainGAN, and SAStainDiff both changed the background colour (PBP = 56–99%) and introduced artificial texture (BG = 1.20–9.43). SAStainDiff produced the strongest textural changes (BG = 9.43), indicating that generative methods can introduce artificial structures into regions that should remain empty. Introducing artificial structures may affect downstream tissue and nuclei detection and nuclei segmentation, compromising the reliability of subsequent analyses. The amount and spatial arrangement of white background within a tile, for example, a tile that is half tissue and half background, or white gaps between tissue regions, also affect normalization performance (see the qualitative comparison of methods in supplementary materials Fig. S2).

### *5.3. Artifact Sensitivity*

The normalization methods fell into three groups based on their sensitivity to artifacts in the image tiles (Fig. 3c): low (StainNet, Pix2Pix, StainGAN; $\Delta E_{00} \leq 6.6$, SSIM ≥ 0.93), moderate (Ruifrok, Macenko, Multi-Macenko, CycleGAN, Vahadane; $\Delta E_{00}$ 8.9–10.3, SSIM 0.88–0.91) and high (Histogram Matching, SAStainDiff, Reinhard; $\Delta E_{00} \geq 10.9$, JSD ≥ 0.39, SSIM ≤ 0.85). The high-sensitivity group includes the two methods that derive their transformation directly from whole-tile statistics (Reinhard and Histogram Matching), so any change in tile content changes the normalization of every pixel. StainNet was the least sensitive compared to other deep learning methods (see the qualitative comparison of methods in supplementary materials Fig. S3). Overall, statistics-based methods are more sensitive to artifacts than deep learning-based methods because their parameter estimates are derived from the entire tile content and can therefore be biased by artifacts. Therefore, artifacts should be removed before normalization, or the normalization method should be designed to handle artifacts, to limit their impact on subsequent analysis.

### *5.4. Impact of Color Normalization on Nuclei Detection and Classification*

Fig. 3d presents detection and classification performance metrics. For a given method, PQ varied strongly across datasets, with all methods performing best on CoNSeP and ILCD and worst on Lizard. For a given dataset, methods differed only slightly. Averaged across datasets, PQ ranged only from 0.33–0.34. No single method was best across datasets. These results indicate that color normalization alone cannot overcome dataset shift, which may arise from factors beyond color variation, including magnification, tissue type, and annotation procedure. Differences in magnification can change the apparent size and morphological characteristics of nuclei, further limiting cross-dataset generalization. Consequently, models trained on multi-organ datasets such as PanNuke may not generalize well to single-organ datasets, and vice versa, particularly when the datasets differ in magnification and tissue composition (Jabar *et al.*, 2026).

Moreover, normalization methods should be designed and evaluated with respect to the downstream task rather than color fidelity, for example through joint training or structure-preserving constraints that prevent generative models from changing nuclei structure. Finally, because the performance varies across datasets, select normalization on target-domain validation data, or combine methods through test-time adaptation, rather than applying a fixed default. **Table 1** summarizes the methods' relative performance across the four analyses. The ranking is meant to identify the best-performing groups of methods rather than a strict order. Here the three groups are presented: the best-performing group, StainNet, StainGAN, and Pix2Pix (mean rank 3.25–4.00); the intermediate group CycleGAN, Vahadane, Multi-Macenko, Histogram Matching, and Ruifrok (4.50–5.50); and the worst performers were Macenko, SAStainDiff, and Reinhard (7.25–9.50). However, no method ranked highest across all analyses; for example, StainNet ranked first in three analyses but tenth in background preservation.

### *5.5. Impact of Input Color Space*

Changing the input color space had only a modest effect on performance ($|\Delta| \leq 0.05$; Fig. 3e), with no color space consistently outperforming the default RGB representation. HLS generally reduced performance across datasets ($\Delta PQ = -0.02$ to $-0.04$), while LUV also tended to reduce performance, particularly on Lizard ($\Delta SQ = -0.05$). LAB and XYZ yielded small improvements on Lizard ($\Delta PQ = +0.02$), but these gains did not generalize to other datasets. HSV and YCbCr had largely negligible effects. In this experiment, the color space was varied only at the model input, without color normalization. Therefore, the effect of color space choice on normalization itself remains unclear. Future normalization methods could take both the color space and nuclei-relevant characteristics into account to improve robustness to staining variability.

### *5.6. Computational Cost*

Processing times were measured on a workstation running Ubuntu 20.04, with a 40-core CPU, 512 GB of RAM and an NVIDIA A100 GPU (80 GB, CUDA 12.2). Classical methods processed one tile at a time, and deep learning methods were run in batches of four tiles. Times are reported as the median of five runs over 100 tiles (**Table 2**). StainNet was the fastest method (3.6 ms per tile). StainGAN and all classical methods except Vahadane required < 20ms per tile, corresponding to about 3 minutes for 10K tiles and 31 minutes or less for 100K tiles. Vahadane (232 ms), Pix2Pix (110 ms) and CycleGAN (88 ms) were considerably slower, requiring 2.4–6.4 hours for 100K tiles. SAStainDiff required 4.1s per tile, corresponding to about 11 hours for 10K tiles and nearly 5 days for 100K tiles, which limits its practicality for large-scale processing.

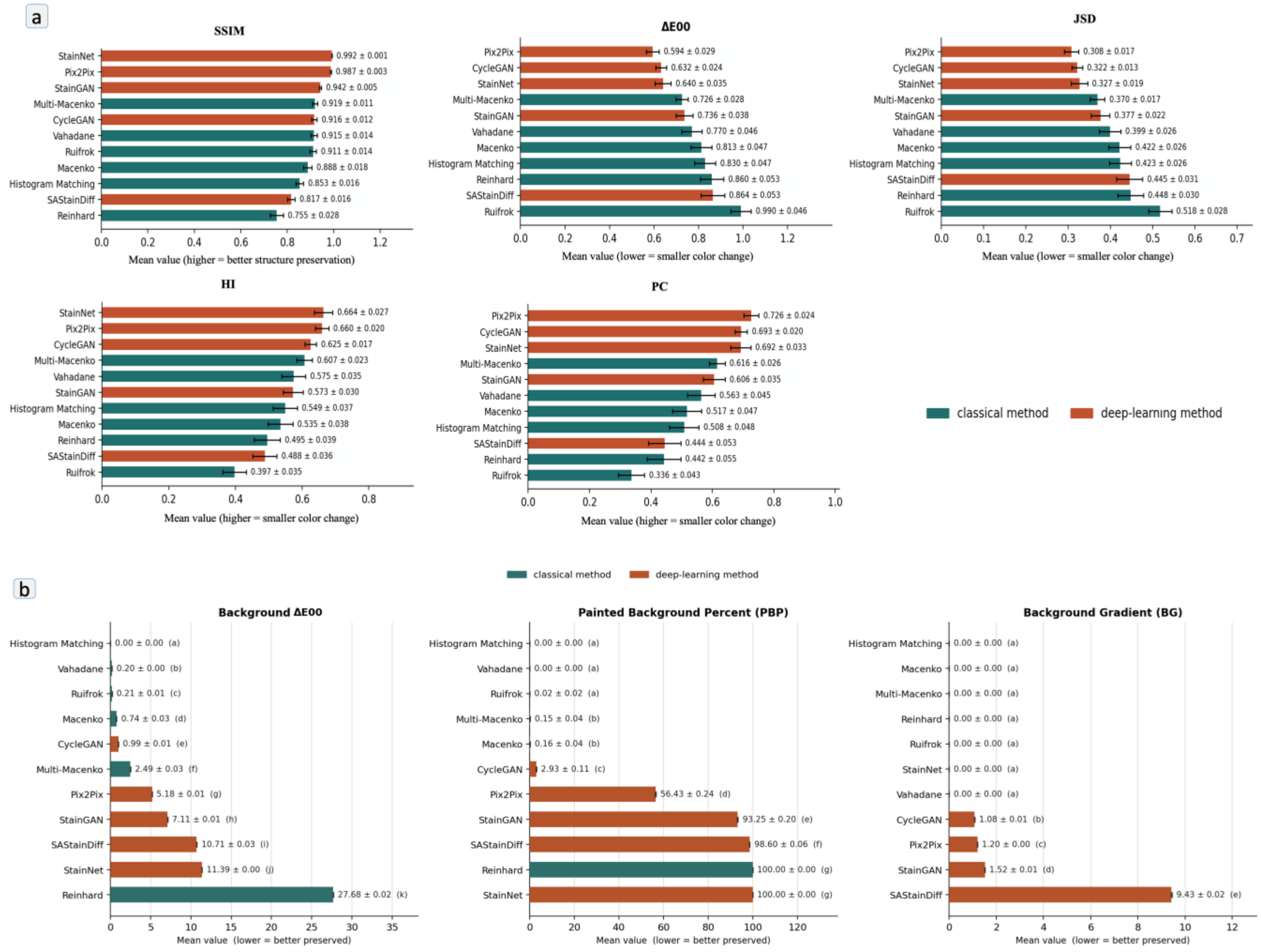

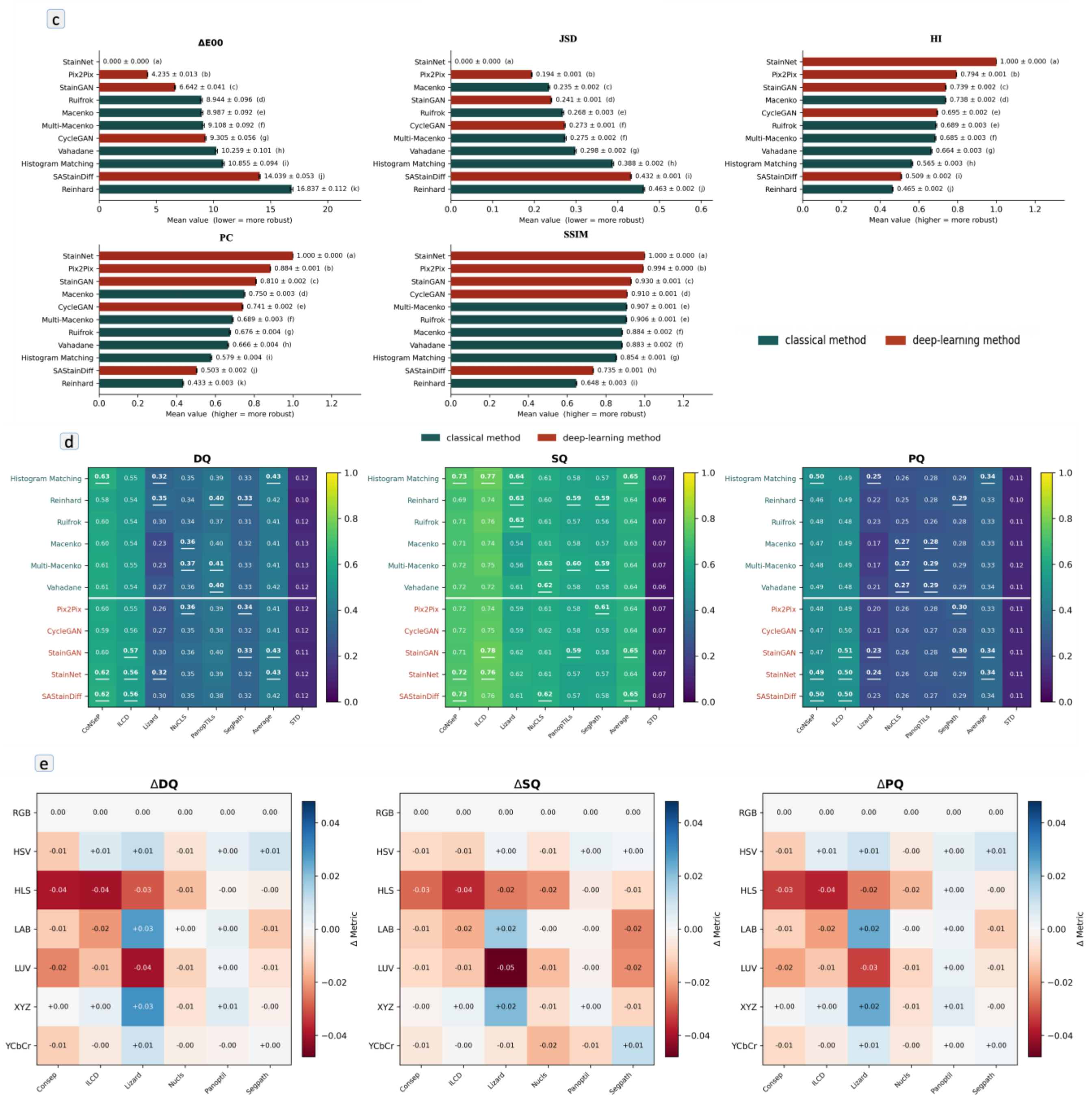

**Fig. 3. Performance metrics:** Evaluation of color normalization methods. (a) Nuclear information preservation: nuclear pixels of each normalized image were compared with the original image using CIEDE2000 colour difference ($\Delta E_{00}$), Jensen–Shannon divergence (JSD), histogram intersection (HI), pearson correlation (PC), and structural similarity index (SSIM), averaged across seven datasets. (b) Background preservation: artifact regions were replaced with white before normalization, and the normalized white regions were assessed using background $\Delta E_{00}$, painted background percentage (PBP), and background gradient (BG). Bars show the mean ± SEM over tiles. Methods sharing a letter are not significantly different (Holm-adjusted paired Wilcoxon signed-rank test, $p \geq 0.05$). (c) Artifact sensitivity: each tile was normalized with and without its artifact, and the two outputs were compared on nuclei pixels within artifact-free tissue. (d) Nuclei detection and classification: one CP-Net model was trained on PanNuke for each normalization method and evaluated on six external test sets using detection quality (DQ), segmentation quality (SQ), and panoptic quality (PQ). The last two columns report the mean and standard deviation across datasets. (e) Effect of input color space: one CP-Net model was trained for each of seven color spaces without normalization on six external test sets using ΔDQ, ΔSQ, and ΔPQ.

**Table 1**. Ranking of color normalization methods across the four analyses (1 = best). (a) SSIM: Nuclei information preservation; (b) $\Delta E_{00}$: background preservation; (c) SSIM: artifact sensitivity; (d) PQ: nuclei detection and classification performance. Mean is the average rank across the four analyses; Overall is the final ranking based on the mean.

| Method | (a) Nuclei, SSIM | (b) Background, $\Delta E_{00}$ | (c) Artifact, SSIM | (d) PQ | Mean | Overall |
|---|---|---|---|---|---|---|
| StainNet | 1 | 10 | 1 | 1 | 3.25 | 1 |
| StainGAN | 3 | 8 | 3 | 1 | 3.75 | 2 |
| Pix2Pix | 2 | 7 | 2 | 5 | 4.00 | 3 |
| CycleGAN | 4 | 5 | 4 | 5 | 4.50 | 4 |
| Vahadane | 4 | 2 | 7 | 5 | 4.50 | 4 |
| M. Macenko | 4 | 6 | 5 | 5 | 5.00 | 6 |
| H. Matching | 9 | 1 | 9 | 1 | 5.00 | 6 |
| Ruifrok | 4 | 3 | 5 | 10 | 5.50 | 8 |
| Macenko | 8 | 4 | 7 | 10 | 7.25 | 9 |
| SAStainDiff | 10 | 9 | 10 | 1 | 7.50 | 10 |
| Reinhard | 11 | 11 | 11 | 5 | 9.50 | 11 |

**Table 2**. Processing time of each color normalization method. Times are the median in ms of five runs over 100 tiles, 10K, and 100K tiles.

| Method | 100 tiles | 10K tiles | 100K tiles |
|---|---|---|---|
| StainNet | 3.6 | 36 s | 6.0 min |
| H. Matching | 7.4 | 1.2 min | 12.3 min |
| StainGAN | 10.6 | 1.8 min | 17.7 min |
| Reinhard | 11.2 | 1.9 min | 18.7 min |
| Macenko | 11.5 | 1.9 min | 19.2 min |
| M.Macenko | 11.6 | 1.9 min | 19.3 min |
| Ruifrok | 18.4 | 3.1 min | 30.7 min |
| CycleGAN | 87.7 | 14.6 min | 2.4 h |
| Pix2Pix | 109.6 | 18.3 min | 3.0 h |
| Vahadane | 231.9 | 38.6 min | 6.4 h |
| SAStainDiff | 4076.4 | 11.3 h | 113 h |

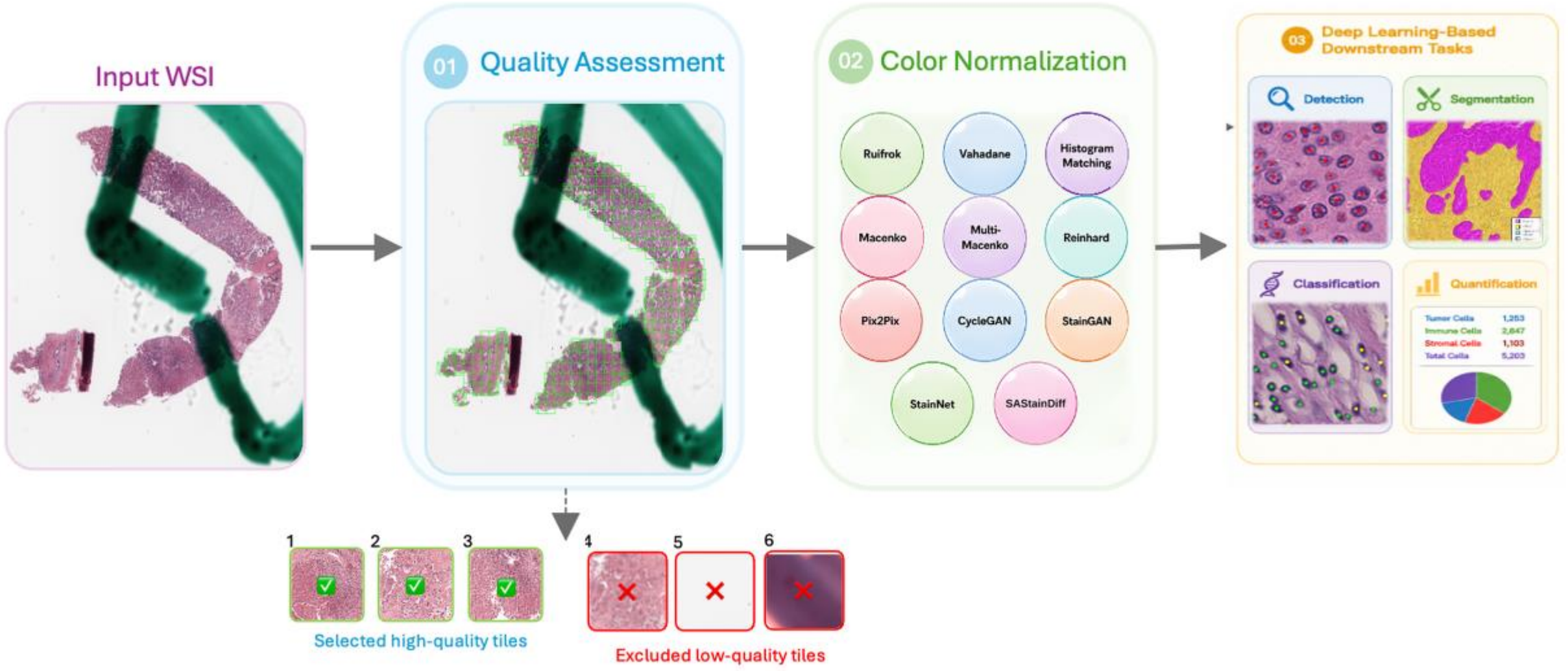

**Fig. 4**. Overview of the WSI-QCN preprocessing pipeline.

## 6. WSI Quality Control and Normalization

Existing color normalization methods do not explicitly address image artifacts and can be sensitive to their presence, potentially compromising normalization quality and downstream analysis. To address this limitation, WSI-QCN was developed as an integrated preprocessing pipeline that combines artifact detection and exclusion with color normalization. It aims to improve the quality and consistency of whole-slide image data and facilitate its use in deep learning-based computational pathology tasks, including model training, testing, and evaluation, as well as applications involving foundation models. An overview of the WSI-QCN pipeline is presented in Fig. 4: (01) WSIs are divided into tiles, and tiles affected by artifacts (e.g., blur, tissue folds, or pen marks) or background are excluded through quality assessment (Jabar *et al.*, 2025); (02) the remaining tiles undergo color normalization using one of 11 classical or deep learning-based methods; (03) the resulting tiles are ready for downstream use in deep learning-based tasks. The pipeline is publicly available on GitHub.

## 7. Conclusion

This study evaluated 11 classical and deep-learning color normalization methods across multiple nuclei datasets, assessing structural preservation, artifact sensitivity, background preservation, downstream performance, and input color space. StainNet, StainGAN, and Pix2Pix performed the best overall, whereas Reinhard, SAStainDiff, and Macenko performed the worst across the evaluated datasets and criteria. Despite differences in image quality, downstream detection and classification performance were primarily dependent on dataset characteristics, with neither color fidelity nor structural similarity reliably predicting performance. Input color space had minimal impact, supporting RGB as a suitable default. We introduce WSI-QCN, a pipeline integrating quality assessment and color normalization, to support whole-slide image analysis. The findings highlight the importance of assessing image quality before normalization method selection through target-domain validation rather than relying on fixed defaults.

Our benchmarking work resulted in the development of the WSI-QCN pipeline, which is designed to generate high-quality, standardized H&E image tiles for downstream deep learning applications. Nevertheless, several limitations remain, highlighting opportunities for further improvement:

- Missing scanner metadata limited assessment of cross-scanner robustness. Future work should explore scanner-aware and scanner-agnostic normalization strategies.
- Differences in tissue type and magnification may have influenced downstream performance, as CP-Net was trained on multi-organ PanNuke data but evaluated on single-organ datasets, sometimes across 40× and 20× magnifications. Future studies should control for these factors and evaluate multiple nuclei detection and classification models to assess the generalizability of the findings.
- Need for task-specific, nuclei-aware color normalization methods that preserve nuclei structure, take the color space into account, are robust to artifacts and background, and produce consistent results across tiles and slides without changing tissue content.
- Color normalization did not substantially reduce inter-dataset performance differences, indicating that color alone does not explain dataset shift. Future work should explore stain augmentation, domain adaptation, and task-driven normalization with structure-preserving constraints.
- Tile-based color normalization methods may normalize neighboring tiles differently and cause visible discontinuities in slide-level results. Future work could explore slide-level color normalization or blending of overlapping tiles.

## Declaration of Competing Interest

Dr. Pinato reported lecture fees from Bayer Healthcare, AstraZeneca, EISAI, Bristol Myers Squibb, Roche, Boston Scientific, and Ipsen; travel support from Bristol Myers Squibb, Roche, Bayer Healthcare, and AstraZeneca; and consulting fees from Accession Therapeutics, MAIA Therapeutics, Bristol Myers Squibb, Mina Therapeutics, Galapagos, Parabilis, Boehringer Ingelheim, Ewopharma, EISAI, Ipsen, Roche, H3 Biomedicine, AstraZeneca, DaVolterra, Mursla, Terumo, Avammune Therapeutics, Starpharma, LiFT Biosciences, and Exact Sciences. All outside the submitted work. Dr Rakaee reported honoraria from AstraZeneca. No other COIs were reported.

## Funding

This research is supported by the Northern Norway Regional Health Authority (grant HNF1771-26). The funders had no role in the design and conduct of the study, nor in the decision to submit the manuscript for publication.

## Declaration of generative AI

During preparation of the first draft of this manuscript, the corresponding author used ChatGPT-4o to check grammar and improve readability. The co-authors then reviewed and edited the manuscript as needed. The corresponding author then performed a final review and editing of the manuscript, taking full responsibility for the content of the published article. Generative AI was not used to write the content.

## Data availability

The annotated image tiles for the ILCD datasets will be made available for research purposes upon request. The remaining datasets used in this study are publicly available: PanNuke, CoNSeP, Lizard, NuCLS, PanopTILs, SegPath, and TCGA artifacts dataset. The source code for the proposed pipeline is publicly available on GitHub.

## References

Gonzalez, R.C. and Woods, R.E. (2018) Digital Image Processing. 4th ed. New York: Pearson. ISBN: 978-0133356724

Hoque, Md.Z. *et al.* (2024) ‘Stain Normalization Methods for Histopathology Image Analysis: A Comprehensive Review and Experimental Comparison’, *Information Fusion*, 102. Available at: https://doi.org/10.1016/j.inffus.2023.101997

Isola, P. *et al.* (2017) ‘Image-to-Image Translation with Conditional Adversarial Networks’, *2017 IEEE Conference on Computer Vision and Pattern Recognition (CVPR)*. Honolulu, USA: IEEE. Available at: https://doi.org/10.1109/CVPR.2017.632

Ivanov, D., Barbano, C.A. and Grangetto, M. (2025) ‘Multi-target Stain Normalization for Histology Slides’, *International Workshop on Medical Optical Imaging and Virtual Microscopy Image Analysis*. Marrakesh, Morocco: Springer-Verlag. Available at: https://doi.org/10.1007/978-3-031-77786-8_4

Jabar, F. *et al.* (2025) ‘Fully Automatic Content-Aware Tiling Pipeline for Pathology Whole Slide Images’, *Intelligence-Based Medicine*, 12(1). Available at: https://doi.org/10.1016/j.ibmed.2025.100318

Jabar, F. *et al.* (2026) ‘CellPrior-net: Prior-Guided Nuclei Detection and Classification for H&E Whole-Slide Images’, *Journal of Pathology Informatics*, 23. Available at: https://doi.org/10.1016/j.jpi.2026.100716

Kang, H. *et al.* (2021) ‘StainNet: A Fast and Robust Stain Normalization Network’, *Frontiers in Medicine*, 8. Available at: https://doi.org/10.3389/fmed.2021.746307

Khan, A.M. *et al.* (2014) 'A Nonlinear Mapping Approach to Stain Normalization in Digital Histopathology Images Using Image-Specific Color Deconvolution', *IEEE Transactions on Biomedical Engineering*, 61(6). Available at: https://doi.org/10.1109/TBME.2014.2303294

Khan, U. *et al.* (2026) 'Staining Normalization in Histopathology: Method Benchmarking Using Multicenter Dataset', *Scientific Reports*, 16(1). Available at: https://doi.org/10.1038/s41598-026-40943-3

Lin, S. *et al.* (2025) 'Impact of Stain Variation and Color Normalization for Prognostic Predictions in Pathology', *Scientific Reports*, 15(1). Available at: https://doi.org/10.1038/s41598-024-83267-w

Macenko, M. *et al.* (2009) 'A Method for Normalizing Histology Slides for Quantitative Analysis', *2009 IEEE International Symposium on Biomedical Imaging: From Nano to Macro*. IEEE. Available at: https://doi.org/10.1109/ISBI.2009.5193250

Reinhard, E. *et al.* (2001) 'Color Transfer Between Images', *IEEE Computer Graphics and Applications*, 21(4). Available at: https://doi.org/10.1109/38.946629

Ruifrok, A.C. and Johnston, D.A. (2001) 'Quantification of Histochemical Staining by Color Deconvolution.', *Analytical and quantitative cytology and histology*, 23(4). PMID: 11531144

Shaban, M.T. *et al.* (2019) 'Staingan: Stain Style Transfer for Digital Histological Images', *2019 IEEE 16th International Symposium on Biomedical Imaging (ISBI 2019)*. Venice, Italy: IEEE. Available at: https://doi.org/10.1109/ISBI.2019.8759152

Tellez, D. *et al.* (2019) 'Quantifying the Effects of Data Augmentation and Stain Color Normalization in Convolutional Neural Networks for Computational Pathology', *Medical Image Analysis*, 58. Available at: https://doi.org/10.1016/j.media.2019.101544

Vahadane, A. *et al.* (2016) 'Structure-Preserving Color Normalization and Sparse Stain Separation for Histological Images', *IEEE Transactions on Medical Imaging*, 35(8). Available at: https://doi.org/10.1109/TMI.2016.2529665

Velastegui, R. and Pedersen, M. (2021) 'The Impact of Using Different Color Spaces in Histological Image Classification using Convolutional Neural Networks', *2021 9th European Workshop on Visual Information Processing (EUVIP)*. Paris, France: IEEE, pp. 1–6. Available at: https://doi.org/10.1109/EUVIP50544.2021.9484035

Xu, C. *et al.* (2025) 'Stain Normalization of Histopathological Images Based on Deep Learning: A Review', *Diagnostics*, 15(8). Available at: https://doi.org/10.3390/diagnostics15081032

Yang, H. *et al.* (2025) 'SAStainDiff: Self-Supervised Stain Normalization by Stain Augmentation Using Denoising Diffusion Probabilistic Models', *Biomedical Signal Processing and Control*, 107(1). Available at: https://doi.org/10.1016/j.bspc.2025.107861

Zhu, J.-Y. *et al.* (2017) 'Unpaired Image-to-Image Translation Using Cycle-Consistent Adversarial Networks', *2017 IEEE International Conference on Computer Vision (ICCV)*. Venice, Italy: IEEE. Available at: https://doi.org/10.1109/ICCV.2017.244

# Supplementary Materials

**ConSeP** (Graham et al., 2019): This dataset consists of 41 image tiles collected from colon cancer patients at the University Hospitals Coventry and Warwickshire, UK. The original tile size is 1000 × 1000 pixels with a spatial resolution of 0.25 µm/px at 40× magnification. The annotations were obtained by a pathologist and subsequently reviewed by a second pathologist to achieve consensus. The annotations were obtained by a pathologist and subsequently reviewed by a second pathologist to achieve consensus: 983 tiles, 40,113 cells, Omnyx VL120 scanner type.

**ILCD** (Falah et al., 2026): This dataset consists of 19,016 image tiles collected from 40 WSIs of liver cancer patients at Imperial College London across multiple centers. Each tile has a size of 256 × 256 pixels with a spatial resolution of 0.25 µm/px at 40× magnification. The annotations were initially performed by two PhD students and one postdoctoral researcher and were subsequently reviewed by a pathologist to ensure annotation quality and accuracy: 19016 tiles, 356312 cells, unknown multicenter scanner type.

**Lizard** (Graham et al., 2021): This dataset extends the CoNSeP dataset by integrating colon cancer tissue images collected from multiple centers in the USA and four hospitals in China. The dataset has a spatial resolution of 0.50 µm/px using 20× magnification and contains 270 image tiles of varying sizes, with an average dimension of 1016 × 917 pixels. The annotations were generated through automatic, semi-automatic, and manual boundary refinement procedures: 7429 tiles, 794334 cells, unknown multicenter scanner type.

**NuCLS** (Amgad et al., 2022): This dataset is a crowdsourced collection of nuclei annotations from breast cancer tissue obtained from TCGA. The dataset was acquired at a spatial resolution of 0.20 µm/px using 20× magnification and contains 151 large image tiles of varying sizes, with an average dimension of 3300 × 2500 pixels. This dataset was annotated by medical students and pathologists: 30464 tiles, 1539799 cells, unknown multicenter scanner type.

**PanopTILs** (Liu et al., 2024): This dataset provides region- and cell-level annotations obtained from breast cancer patients within the TCGA cohort. The dataset was acquired at a spatial resolution of 0.25 µm/px using 40× magnification and consists of 157 large image tiles, each 1024 × 1024 pixels. It includes pathologist-approved nuclei annotations for multiple cell types: 1743 tiles, 52355 cells, unknown multicenter scanner type.

**SegPath** (Komura et al., 2023): This dataset consists of 158,687 registered H&E and immunofluorescence (IF) image patches collected from 18 organs at the University of Tokyo Hospital. The images have a size of 984 × 984 pixels with a spatial resolution of 0.22 µm/px at 40× magnification. Annotations were automatically generated using registered H&E and IF images, together with region-wise semantic segmentation masks derived from IF staining: 542238 tiles, 3631088 cells, Hamamatsu Nanozoomer S60 scanner type.

**PanNuke** (Jevgenij et al., 2019): This dataset consists of 7904 image tiles collected from 19 different organs. The images have a size of 256 × 256 pixels with a spatial resolution of 0.25 µm/px at 40× magnification. Annotations were semi-automatically generated and validated by expert pathologists: 7904 tiles, 162971 cells, unknown multicenter scanner type.

**TCGA artifacts dataset** (Jabar et al., 2025): This dataset includes 18H&E-stained WSIs, which had artifacts, including six different organs: brain, breast, bladder, lung, liver and kidney, obtained from the TCGA cohort. Each organ is represented by three WSIs (3 WSIs per organ, totaling 18). The brain and bladder WSIs were scanned at 0.5 µm/px (20× magnification), while the remaining organs were scanned at 0.25 µm/px (40× magnification). In total, 10,701 tiles were generated, with the following distribution: 3117 artifact-free (qualified tissue) tiles, 1350 tiles with folding, and 6243 tiles with blurring artifacts.

**Table S1**. Overview of traditional and deep learning-based color normalization methods for H&E whole-slide images, summarizing their approaches, strengths, and limitations.

| Method | Approach | Strengths | Limitations |
|---|---|---|---|
| **Histogram Matching** Gonzalez et al., 2018); GitHub | Traditional – color statistics: Matches the full histogram of each RGB channel to the reference | Simple, fast, no training; matches the whole distribution | Ignores stains and spatial structure; RGB channels matched independently; sensitive to reference choice and artifacts |
| **Reinhard** Reinhard et al., 2001); GitHub | Traditional – color statistics: Matches per-channel mean and standard deviation in a decorrelated LAB (lαβ) color space | fast and simple; no training; widely used; available in standard libraries | One global transform for all pixels; ignores H&E stains; sensitive to background and artifacts |
| **Ruifrok** Ruifrok et al., 2001; GitHub | Traditional – stain separation: OD color deconvolution with a predefined stain matrix; rescales H and E concentrations to the reference | Fast; no stain estimation; no training; produces separate H and E channels | Fixed stain matrix does not adapt to slide-to-slide variation; matches only stain intensity, not hue; sensitive to artifacts |
| **Macenko** Macenko et al., 2009; GitHub | Traditional – stain separation: Estimates image-specific H and E vectors by SVD of OD values; reconstructs with the reference stain matrix | Adaptive stain estimation; fast; no training; widely used and available in standard libraries | Single reference; assumes two stains; unstable when one stain dominates or vectors overlap; sensitive to background and artifacts |
| **Vahadane** Vahadane et al., 2016; GitHub | Traditional – stain separation: Sparse non-negative matrix factorization (SNMF) for stain separation; combines source concentrations with the reference stain basis | Preserves tissue structure; no training; widely used and available in standard libraries | Slow; single reference; sensitive to initialization; sensitive to artifacts |
| **Multi-Target Macenko** Ivanov et al., 2025; GitHub | Traditional – stain separation: Macenko with the reference stain matrix averaged over several reference images | Fast; less dependent on a single reference; parameter-free; better generalization to external data | Small gains over Macenko; sensitive to background and artifacts |
| **Pix2pix** (Isola et al., 2017); GitHub | Deep learning – paired GAN: Conditional GAN (U-Net or DenseUNet generator) that translates grayscale input into the target stain appearance | Preserves tissue structure; learns a non-linear, context-aware mapping; no reference needed; robust against artifacts | Requires paired training data; may hallucinate structures or desaturate output; sensitive to background; slow inference |
| **CycleGan** (Zhu et al., 2017; GitHub | Deep learning – unpaired GAN: Unpaired translation between stain domains with adversarial and cycle-consistency losses | No paired data needed; learns the target distribution from a whole dataset; preserves tissue structure | Costly training; may introduce artifacts or alter nuclear texture; needs one model per domain pair; slow inference; sensitive to background |
| **StainGan** (Shaban et al., 2019); GitHub | Deep learning – unpaired GAN: CycleGAN applied to stain transfer between scanners or labs | No expert reference selection; outperformed traditional methods; fast inference | Evaluated on two domains only; GAN artifacts; slower than StainNet; sensitive to background |
| **StainNet** (Kang et al., 2021); GitHub | Deep learning – distilled network: Distills StainGAN into a lightweight network applied pixel by pixel | Faster than StainGAN; preserves tissue structure; robust against artifacts; fast inference | Inherits teacher biases; requires a trained GAN; sensitive to background |
| **SAStainDiff** (Yang et al., 2025); GitHub | Deep learning – diffusion: Denoising diffusion model generating normalized images through iterative denoising | Learns a non-linear, context-aware mapping; no reference needed; | Sensitive to background and artifacts; slow inference |

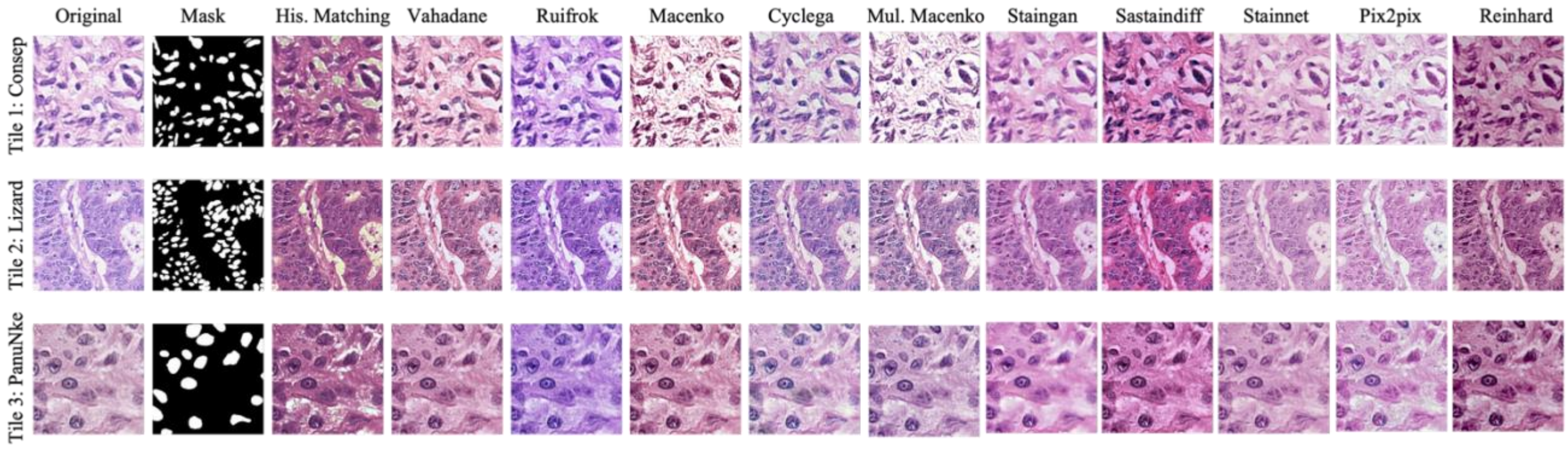

**Fig. S1. Nuclei structure preservation**: Visual comparison of color normalization methods across three H&E-stained histopathology image tiles. Representative tiles from CoNSeP (Tile 1), Lizard (Tile 2), and PanNuke (Tile 3) are shown before normalization (Original), alongside their corresponding nuclei masks (Mask) and results from 11 color normalization methods.

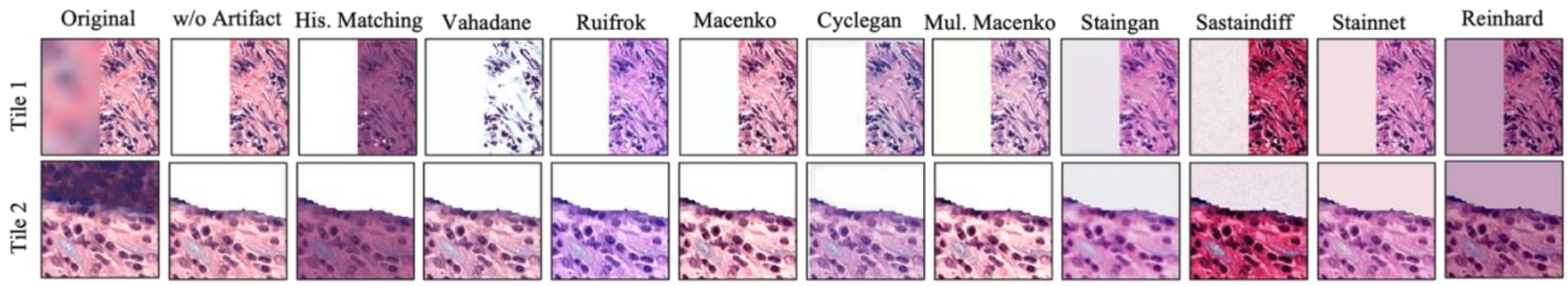

**Fig. S2. Background preservation**: Visual comparison of color normalization methods in H&E-stained histopathology images. Representative image tiles illustrate the original images, images after artifact removal (w/o Artifact) replaced with near-white, and the results of 11 color normalization methods.

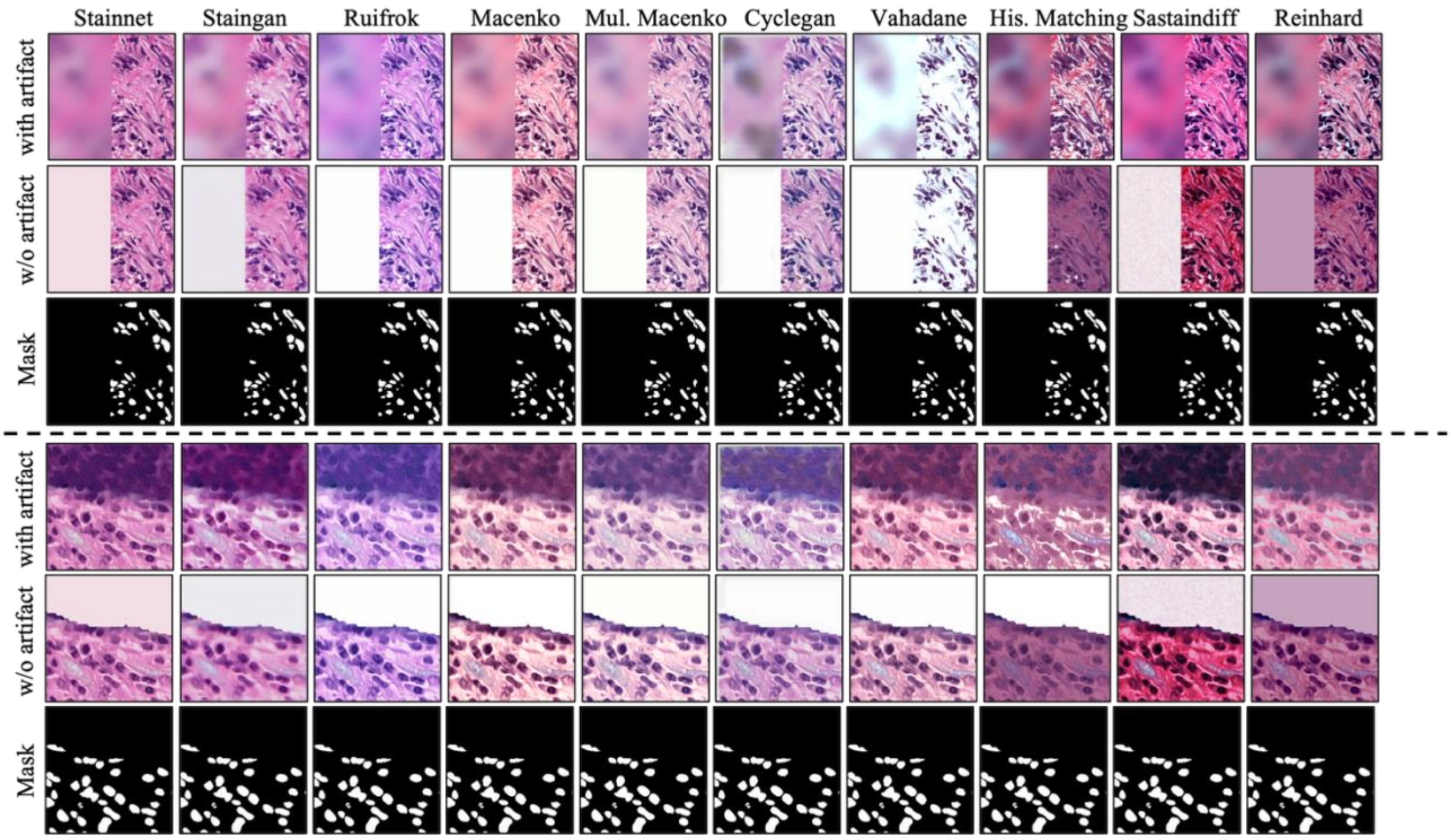

**Fig. S3. Artifact sensitivity**: Visual examples of artifact sensitivity of stain-normalization methods. Two representative tiles are shown, one with a blur artifact (rows 1–3) and one with a tissue fold (rows 4–6). Rows 1 and 4: normalized tile with the artifact present. Rows 2 and 5: the same tile normalized after the artifact region was removed (replaced by near-white). A method that is robust to artifacts produces the same colors in this region whether or not the artifact was present.

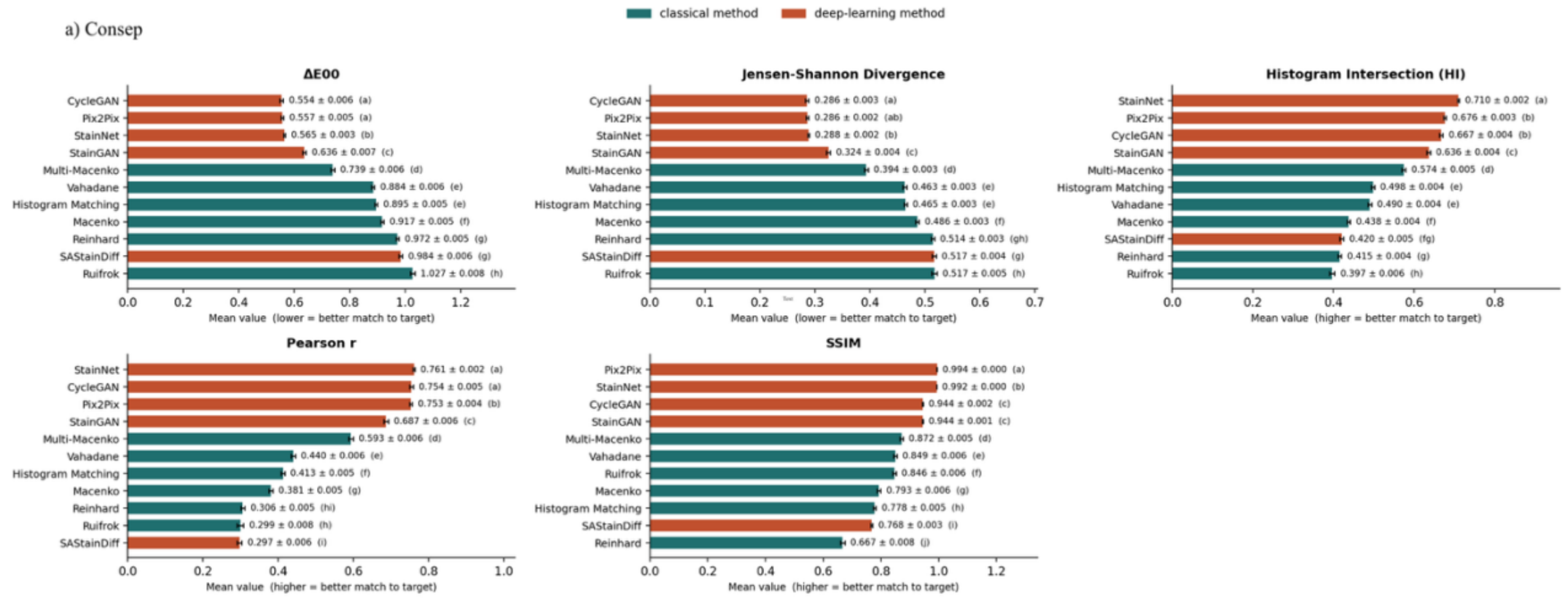

b) Segpath

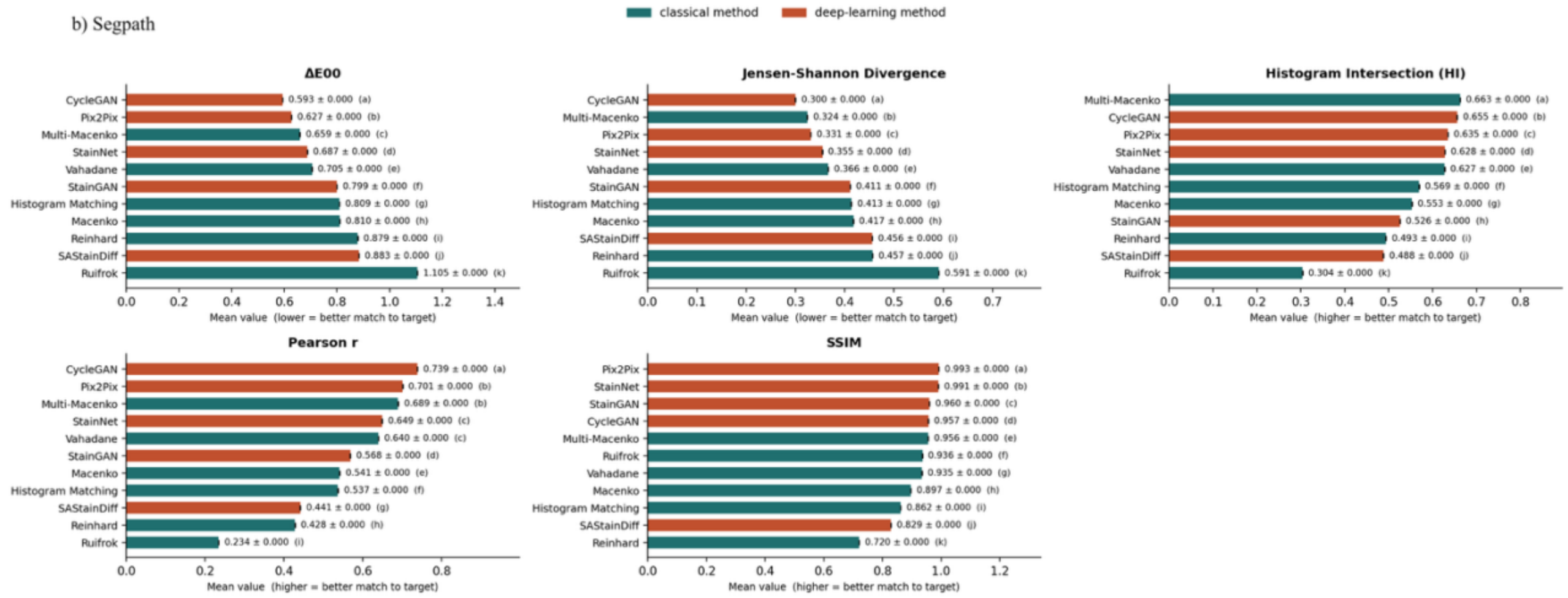

c) Panoptils

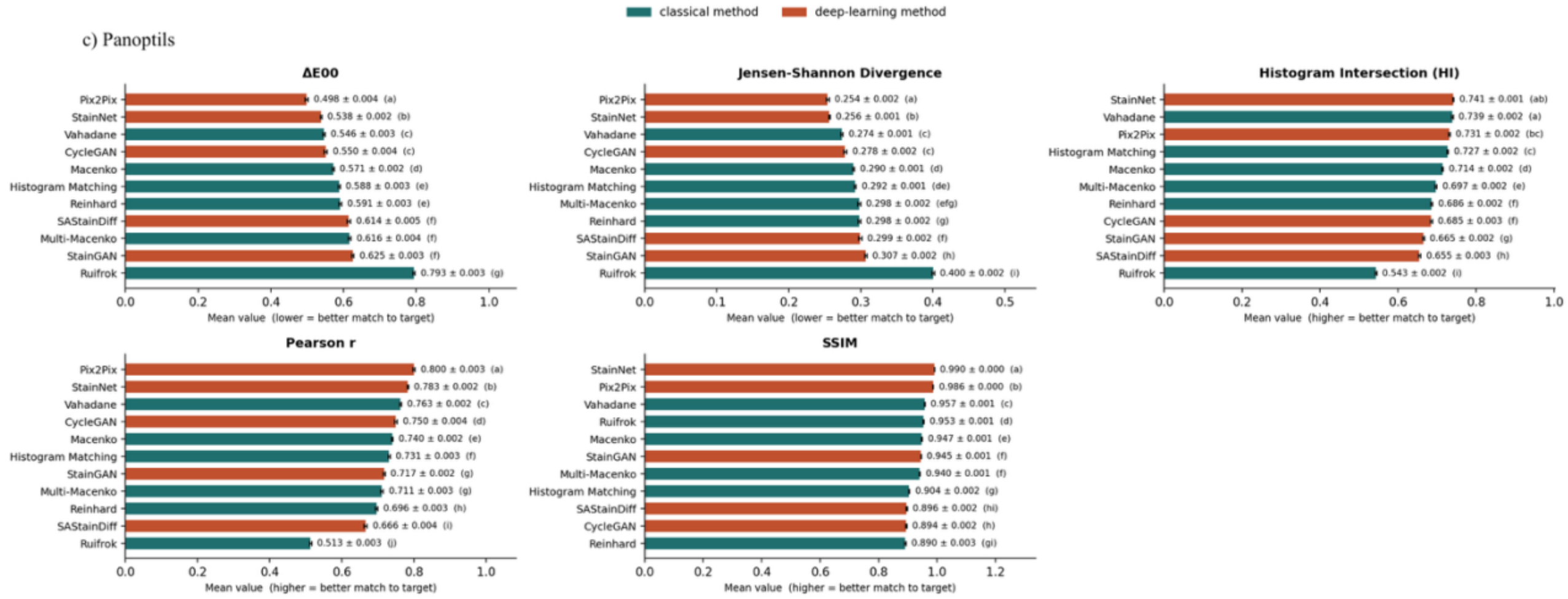

d) PanNuke

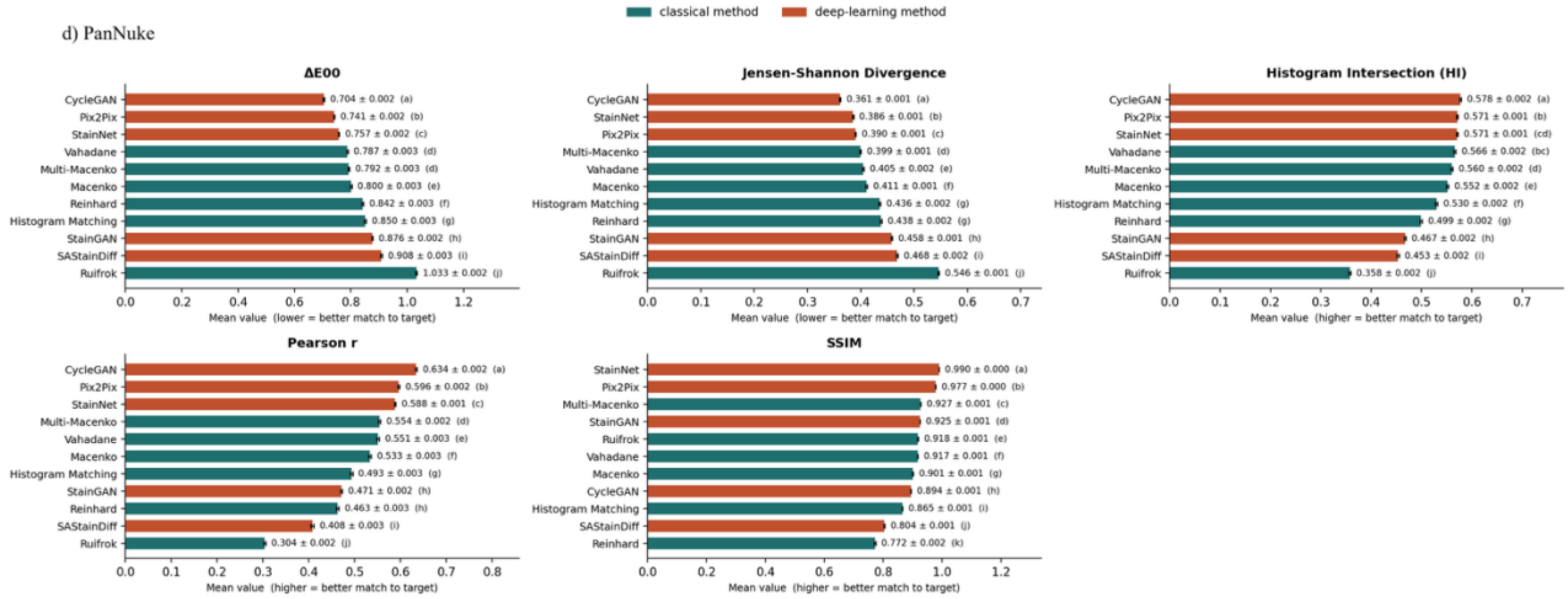

e) Nucls

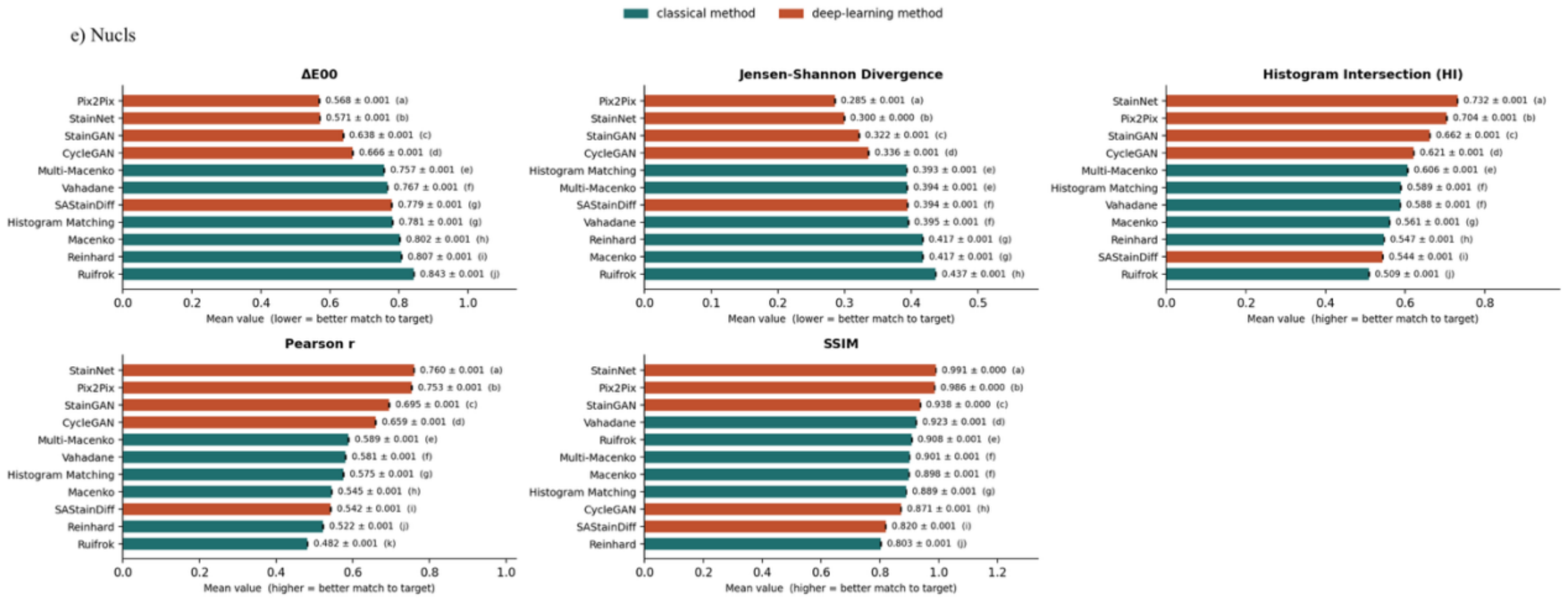

f) Lizard

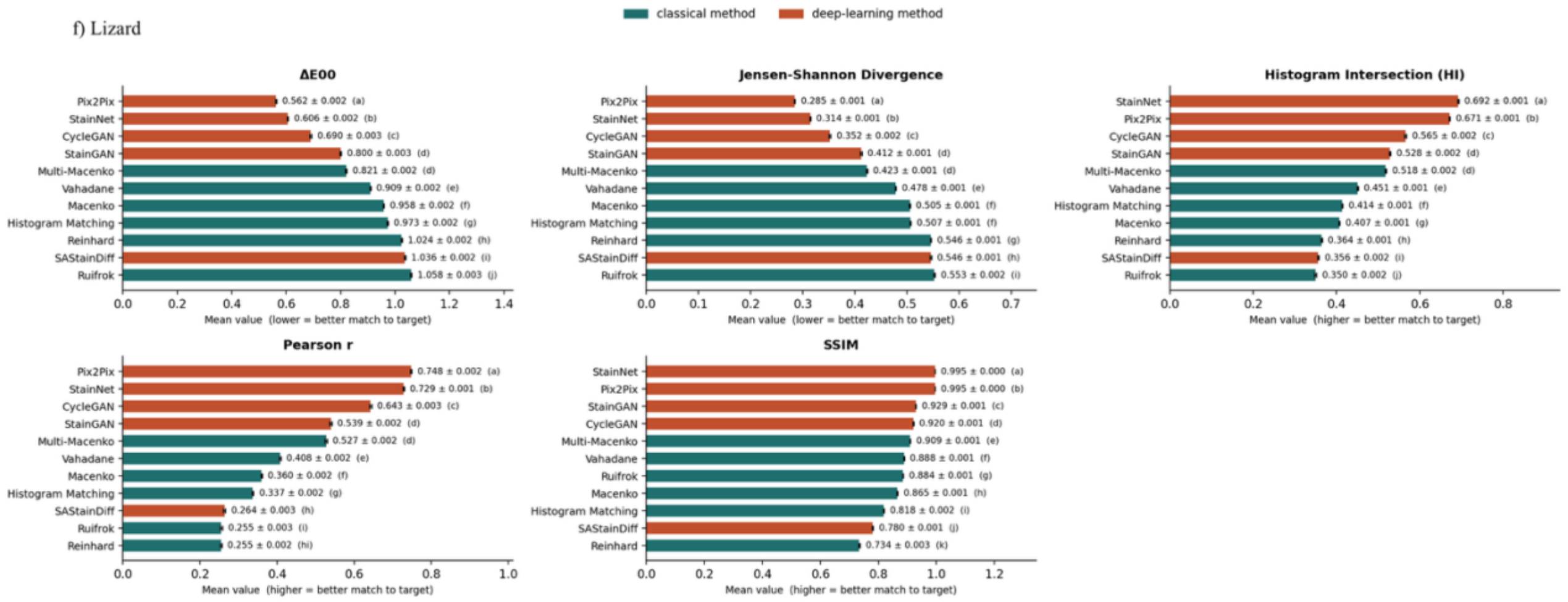

g) ILCD

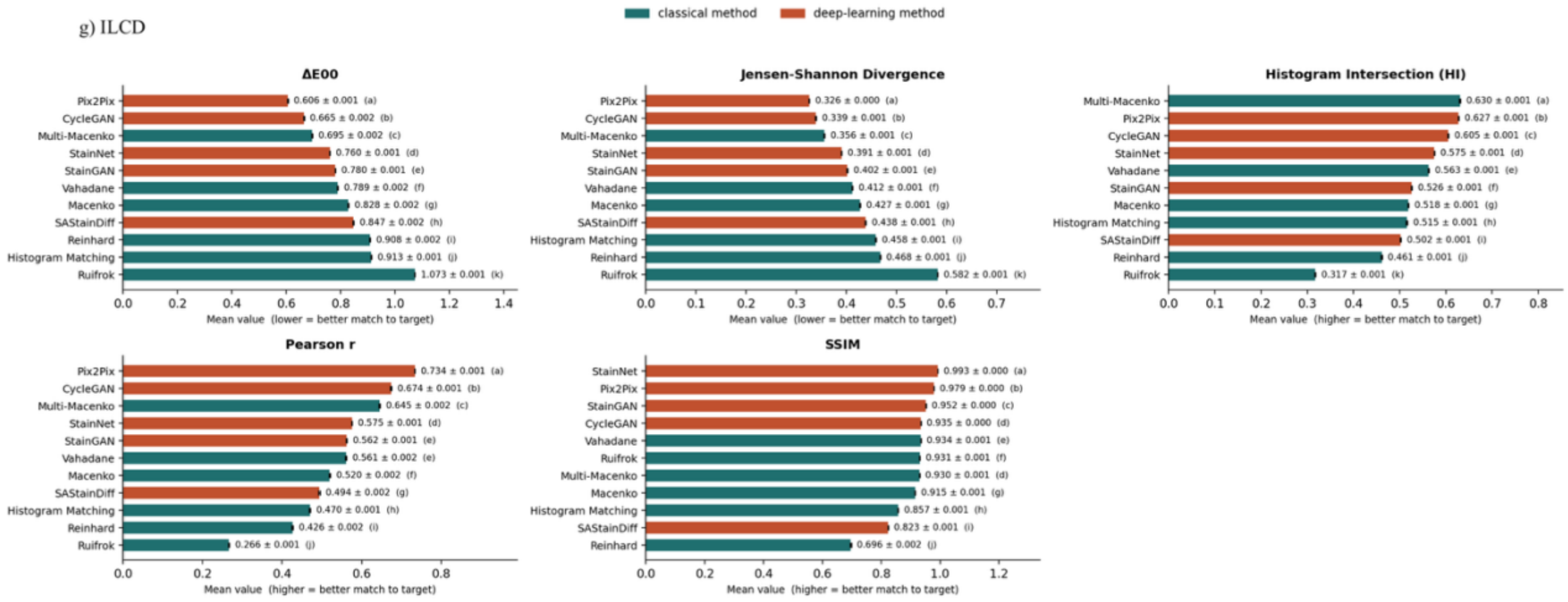

**Fig. S4**. Nuclear structure preservation after color normalization, shown separately for each dataset (panels a–g; e.g. g, ILCD). Nuclear pixels of each normalized image were compared with those of the original image. SSIM measures structural preservation; $\Delta E_{00}$, Jensen–Shannon divergence (JSD), histogram intersection (HI) and Pearson correlation (PC) quantify the magnitude of the color change. Bars show the mean ± SEM over tiles. Methods sharing a letter are not significantly different (Holm-adjusted paired Wilcoxon signed-rank test, $p \geq 0.05$).

## References

Amgad, M., et al. (2022). NuCLS: A scalable Crowdsourcing Approach and Dataset for Nucleus Classification and Segmentation in Breast Cancer. *GigaScience*, *11*(1). https://doi.org/10.1093/gigascience/giac037

Gonzalez, R. C., & Woods, R. E. (2018). *Digital Image Processing* (3rd ed., Vol. 1). Pearson. https://doi.org/ISBN:978-1-292-22304-9

Graham, S., et al. (2021, October 11). Lizard: A Large-Scale Dataset for Colonic Nuclear Instance Segmentation and Classification. *2021 IEEE/CVF International Conference on Computer Vision Workshops*. https://doi.org/10.1109/ICCVW54120.2021.00082

Graham, S., et al. (2019). Hover-Net: Simultaneous Segmentation and Classification of Nuclei in Multi-tissue Histology Images. *Medical Image Analysis*, *58*(1). https://doi.org/10.1016/j.media.2019.101563

Isola, P., Zhu, J.-Y., Zhou, T., & Efros, A. A. (2017, July). Image-to-Image Translation with Conditional Adversarial Networks. *2017 IEEE Conference on Computer Vision and Pattern Recognition (CVPR)*. https://doi.org/10.1109/CVPR.2017.632

Ivanov, D., Barbano, C. A., & Grangetto, M. (2025, October). Multi-target Stain Normalization for Histology Slides. *International Workshop on Medical Optical Imaging and Virtual Microscopy Image Analysis*. https://doi.org/10.1007/978-3-031-77786-8_4

Jabar, F., et al. (2026). CellPrior-net: Prior-Guided Nuclei Detection and Classification for H&E Whole-Slide Images. *Journal of Pathology Informatics*, *23*. https://doi.org/10.1016/j.jpi.2026.100716

Jabar, F., et al. (2025). Fully Automatic Content-Aware Tiling Pipeline for Pathology Whole Slide Images. *Intelligence-Based Medicine*, *12*(1). https://doi.org/10.1016/j.ibmed.2025.100318

Jevgenij, G., Navid Alemi, K., Ksenija, B., Ali, K., & Nasir, R. (2019). PanNuke: An Open Pan-cancer Histology Dataset for Nuclei Instance Segmentation and Classification. *European Congress on Digital Pathology*. https://doi.prg/10.1007/978-3-030-23937-4_2

Kang, H., Luo, D., Feng, W., Zeng, S., Quan, T., Hu, J., & Liu, X. (2021). StainNet: A Fast and Robust Stain Normalization Network. *Frontiers in Medicine*, *8*. https://doi.org/10.3389/fmed.2021.746307

Komura, D., et al. (2023). Restaining-Based Annotation for Cancer Histology Segmentation to Overcome Annotation-Related Limitations Among Pathologists. *Patterns*, *4*(2). https://doi.org/10.1016/j.patter.2023.100688

Liu, S., Amgad, M., More, D., Rathore, M. A., Salgado, R., & Cooper, L. A. D. (2024). A Panoptic Segmentation Dataset and Deep-Learning Approach for Explainable Scoring of Tumor-Infiltrating Lymphocytes. *Npj Breast Cancer*, *10*(1). https://doi.org/10.1038/s41523-024-00663-1

Macenko, M., et al. (2009, June). A Method for Normalizing Histology Slides for Quantitative Analysis. *2009 IEEE International Symposium on Biomedical Imaging: From Nano to Macro*. https://doi.org/10.1109/ISBI.2009.5193250

Reinhard, E., Adhikhmin, M., Gooch, B., & Shirley, P. (2001). Color Transfer Between Images. *IEEE Computer Graphics and Applications*, *21*(4). https://doi.org/10.1109/38.946629

Ruifrok, A. C., & Johnston, D. A. (2001). Quantification of Histochemical Staining by Color Deconvolution. *Analytical and Quantitative Cytology and Histology*, *23*(4). PMID: 11531144

Shaban, M. T., Baur, C., Navab, N., & Albarqouni, S. (2019, April). Staingan: Stain Style Transfer for Digital Histological Images. *2019 IEEE 16th International Symposium on Biomedical Imaging (ISBI 2019)*. https://doi.org/10.1109/ISBI.2019.8759152

Vahadane, A., et al. (2016). Structure-Preserving Color Normalization and Sparse Stain Separation for Histological Images. *IEEE Transactions on Medical Imaging*, *35*(8). https://doi.org/10.1109/TMI.2016.2529665

Yang, H., Lyu, M., Yan, S., Zhong, T., Li, J., Xu, T., Xie, H., & Liu, S. (2025). SAStainDiff: Self-Supervised Stain Normalization by Stain Augmentation Using Denoising Diffusion Probabilistic Models. *Biomedical Signal Processing and Control*, *107*(1). https://doi.org/10.1016/j.bspc.2025.107861

Zhu, J.-Y., Park, T., Isola, P., & Efros, A. A. (2017, October). Unpaired Image-to-Image Translation Using Cycle-Consistent Adversarial Networks. *2017 IEEE International Conference on Computer Vision (ICCV)*. https://doi.org/10.1109/ICCV.2017.244